\documentclass[letterpaper, 10 pt, conference]{ieeeconf}  

\IEEEoverridecommandlockouts                              

\usepackage[backend=biber,style=numeric-comp,sorting=none,firstinits=true,maxbibnames=10]{biblatex}
\usepackage{times}
\usepackage{multicol}
\usepackage{hyperref}
\usepackage{graphicx} 
\usepackage{booktabs} 
\usepackage{subfig} 
\usepackage{amsmath, amsfonts}
\usepackage{amssymb}
\usepackage{float}
\usepackage{dblfloatfix}    
\usepackage{flushend} 

\graphicspath{ {figs/} }

\newcommand{\diag}{\mathop{\mathrm{diag}}}

\hypersetup{pdfinfo={
   Title={CALC2.0: Combining Appearance, Semantic and Geometric Information for Robust and Efficient Visual Loop Closure}, 
   Author={Nate Merrill and Guoquan Huang},
   Subject={Visual Place Recognition},
   Keywords={SLAM; Deep learning; Visual place recognition; Autoencoder; Semantic Segmentation; Descriptor; Features; CNN; ConvNet; Feature embedding; VAE},
}}

\begin{document}

\title{ \LARGE \bf 
CALC2.0: Combining Appearance, Semantic and Geometric Information for Robust and Efficient Visual Loop Closure}

\author{Nathaniel Merrill and Guoquan Huang 
\thanks{This work was partially supported by the University of Delaware (UD) College of Engineering, 
the DTRA (HDTRA1-16-1-0039),
and Google Daydream.
} 
\thanks{The authors are with the Dept. of Computer and Information Sciences and Dept. of Mechanical Engineering, University of Delaware, Newark, DE 19716. Email: \tt{\{nmerrill,ghuang\}@udel.edu} }
}

\maketitle

\begin{abstract}
Traditional attempts for loop closure detection typically use hand-crafted features, 
relying on geometric and visual information only, whereas more modern approaches tend to use semantic, appearance or geometric features extracted from deep convolutional neural networks (CNNs).
While these approaches are successful in many applications, they do not utilize all of the information that a monocular image provides, and many of them, particularly the deep-learning based methods, require user-chosen thresholding to actually close loops -- which may impact generality in practical applications.
In this work, we address these issues by extracting all three modes of information from a custom deep CNN  trained specifically for the task of place recognition.
Our network is built upon a combination of a semantic segmentator, Variational Autoencoder (VAE) and triplet embedding network.
The network is trained to construct a global feature space to describe both the visual appearance and semantic layout of an image.
Then local keypoints are extracted from maximally-activated regions of low-level convolutional feature maps, and keypoint descriptors are extracted from these feature maps in a novel way that incorporates ideas from successful hand-crafted features.
These keypoints are matched globally for  loop closure candidates, and then used as a final geometric check to refute false positives.
As a result, the proposed loop closure detection system requires no touchy thresholding,
and is highly robust to false positives -- achieving better precision-recall curves than the state-of-the-art NetVLAD, and with real-time speeds.
\end{abstract}


\section{Introduction}

Visual loop closure (or place recognition) is a key component of long-term SLAM systems; however, it is a difficult task in the general case due to appearance and viewpoint changes.
A long-term autonomous system revisiting a location is subject to extreme variations in viewpoint, weather, and brightness in addition to the possibility of moving objects -- making the use of images for place recognition extremely difficult in practice.
While other sensors such as LiDAR may be able to overcome some of these issues more easily, it is preferable to only carry a camera on small mobile robots such as aerial vehicles.

Traditional loop closure methods have relied on hand-crafted features, typically in tandem with a binary word tree~\cite{GalvezTRO12}.
While these methods are reliable in many cases, large viewpoint and appearance changes can cause false positives.
Features from off-the-shelf convolutional neural networks (CNNs) have shown to outperform hand-crafted features in the loop closure task~\cite{sunderhauf_performance_2015}.
Much recent work has been done to better utilize such CNN features, such as LoST \cite{garg2018lost}, which creates descriptors from dense semantic segmentations and low-level convolution layers to solve opposite-viewpoint place recognition.
Furthermore, others have trained CNNs specifically for the place recognition task.
For example,  NetVLAD \cite{Arandjelovic16} learns to regress and assign VLAD descriptors and cluster centers from image triplets, 
and our recent CALC network~\cite{Merrill2018RSS} utilizes the HOG descriptor \cite{HOG} and random data augmentations to learn an image descriptor in an unsupervised manner.

In this work, building upon our previous method~\cite{Merrill2018RSS}, we seek to learn a novel image representation that incorporates appearance, semantic and geometric information to close loops.
In particular, we construct a novel multi-decoder variational autoencoder (VAE) that forces different characteristics of an image, visual appearance and semantics, into different feature maps of the latent space.
The concept of binning features in this manner is similar to NetVLAD's VLAD core layer~\cite{Arandjelovic16}, but we do not rely on softmax assignment of features since the network assigns them by design.
Furthermore, we utilize the VLAD residual aggregation and normalization scheme to increase the discriminative abilities of our descriptor while allowing efficient comparison by inner product in addition to learning cluster centers for the local descriptors.

Our descriptor consists of local features pertaining to visual appearance and semantic classes, which, combined, creates a discriminative image representation.
To further robustify our method, we extract keypoints and descriptors from maximally activated locations of the network's convolutional feature maps in order to further filter loop closure candidates.
Additionally, our model is forced to discriminate between positive and negative images without the need to label them, since the positives are obtained by random homographies, brightness alteration, and left-right flipping while the negatives are mined from the current training batch.
The result is an efficient yet effective loop closure system which is able to elucidate multiple modes of image information to consistently recognize places.

Specifically, the main contributions of this work include:
\begin{itemize}
    \item We design and construct a novel multi-decoder VAE-based network for robust loop closure, which utilizes visual appearance, semantic and geometric information and
    is shown to outperform the state-of-the-art NetVLAD~\cite{Arandjelovic16} in some experiments. 
    
    \item We develop a novel image descriptor composed of local features encoding visual appearance and semantic information, as well as  
    a new keypoint description method using residual activations of different cells in convolutional feature maps. 
    To better benefit our community, we release our open source implementation at: \url{https://github.com/rpng/calc2.0}.
\end{itemize}

The rest of this paper is organized as follows: 
We discuss related methods that have influenced this work in next section.
In Section~\ref{sec:methodology} we outline the proposed approach to loop closure detection.
In Section~\ref{sec:experiments} we rigorously test the proposed method against the state-of-the-art approach.
Finally,  we conclude the paper and discuss future research directions in Section~\ref{sec:conclusions}.


\section{Related Work} \label{sec:rel_work}

Visual place recognition remains one of the most challenging problems in long-term SLAM, and has been an active research area in recent years.
Many systems have come forth~\cite{Lowry2016TRO}, each with their own benefits, in terms of accuracy, efficiency, storage requirements, or a mix of the three.

One of the most popular loop closure systems is DBoW~\cite{GalvezTRO12}.
This algorithm is based on Bag of Words (BoW) and stores binary features in an efficient vocabulary tree structure for nearest neighbor searches.
Additionally, geometric checks are performed with loop closure candidates as a final step to avoid false positives.
To do so, the binary features are matched for database and query images, and a loop is considered detected if a valid fundamental matrix can be computed.
This method is highly successful in real-world applications, and has been deployed in many SLAM systems such as ORB-SLAM~\cite{murTRO2015,murORB2}.

The recent work of LoST~\cite{garg2018lost} constructs global image descriptors from semantic segmentation predictions. 
Additionally, the LoST-X variant (from the same publication) detects keypoints from maximally activated regions of the low-level {\em conv5} feature maps, and matches them between database and query images using a mix of the Euclidean distance metric and semantic consistency.
This method is highly robust to appearance changes, and is the first method to achieve a meaningful recall rate for opposite-viewpoint place recognition.
However, LoST-X is designed {\textit only} for opposite-viewpoint place recognition due to the left-right flipping of x locations in the keypoint matching algorithm.

NetVLAD~\cite{Arandjelovic16} is widely considered a state-of-the-art place recognition system.
This method constructs a global image descriptor by assigning local descriptors (flattened convolutional feature maps) to different local learned clusters -- similar to traditional VLAD~\cite{Arandjelovic2013}.
NetVLAD requires image triplets to train -- two ``weak'' positive images, and a hard-mined negative image.
By relying on GPS data for place supervision, it is possible to have two positives that do not actually have any visual overlap.
To account for this issue, the authors mine for hard negatives in order to make sure the model does not lose any discriminative power.
NetVLAD is a highly generalizable system that achieves state-of-the-art accuracy; however, it does not provide any method to actually close loops.
It is assumed that a thresholding scheme is to be used based on precision-recall experiments.

The method proposed in~\cite{Schonberger2018semantic} encodes semantic information in a global descriptor by placing partially-observed semantic voxel grids in a VAE.
The VAE is forced to reconstruct the fully-observed semantic subvolume from the latent space.
This method proves to be very accurate in the loop closure task, even for large viewpoint and appearance changes; however, it requires a dense semantic segmentation and depth map, which are expensive to compute.

In our previous work~\cite{Merrill2018RSS}, we have developed a loop closure system dubbed Convolutional Autoencoder for Loop Closure (CALC) that learns geometric information from HOG descriptors.
This method is shown to outperform DBoW among other methods in place recognition, and achieves state-of-the-art extraction and query speeds.
While this method is effective for actually closing loops on the KITTI dataset~\cite{Geiger2012CVPR},
CALC does not provide any method for geometric checks, and, again, relies on thresholding to determine a true positive,
which is part of what we want to improve in this work.

In particular, the proposed CALC2.0 aims to combine the advantages of the aforementioned approaches.
That is, we integrate semantic and appearance information to construct a robust whole-image descriptor -- both through autoencoding as in~\cite{Merrill2018RSS} and triplet embedding as in~\cite{Arandjelovic16}.
The learned features are agreggated by taking the residual from learned cluster centers, and intra-normalized and globally normalized as in~\cite{Arandjelovic16} in order to reduce bursts in the descriptors, and to allow comparison by inner product.
Additionally, we use a keypoint detection scheme that is similar to~\cite{garg2018lost}, 
but a keypoint descriptor that is more similar in concept to hand-crafted descriptors such as BRIEF~\cite{BRIEF} --
replacing smoothed image intensity regions with regions of convolutional feature maps.

\section{CALC2.0: The Proposed Loop Closure Approach} \label{sec:methodology}

In this section, we present in detail the proposed deep loop closure detection system, which is termed CALC2.0 as it is built upon our prior work~\cite{Merrill2018RSS}.
The key idea of the proposed approach is to utilize all modes of information that monocular images offer, including visual appearance, semantics and geometric consistency,
thus accurately and robustly detecting loops while avoiding false positives without the need for user-defined thresholding.
%

\subsection{Network Design}

\begin{figure*} 
	\centering
	\includegraphics[width=\textwidth]{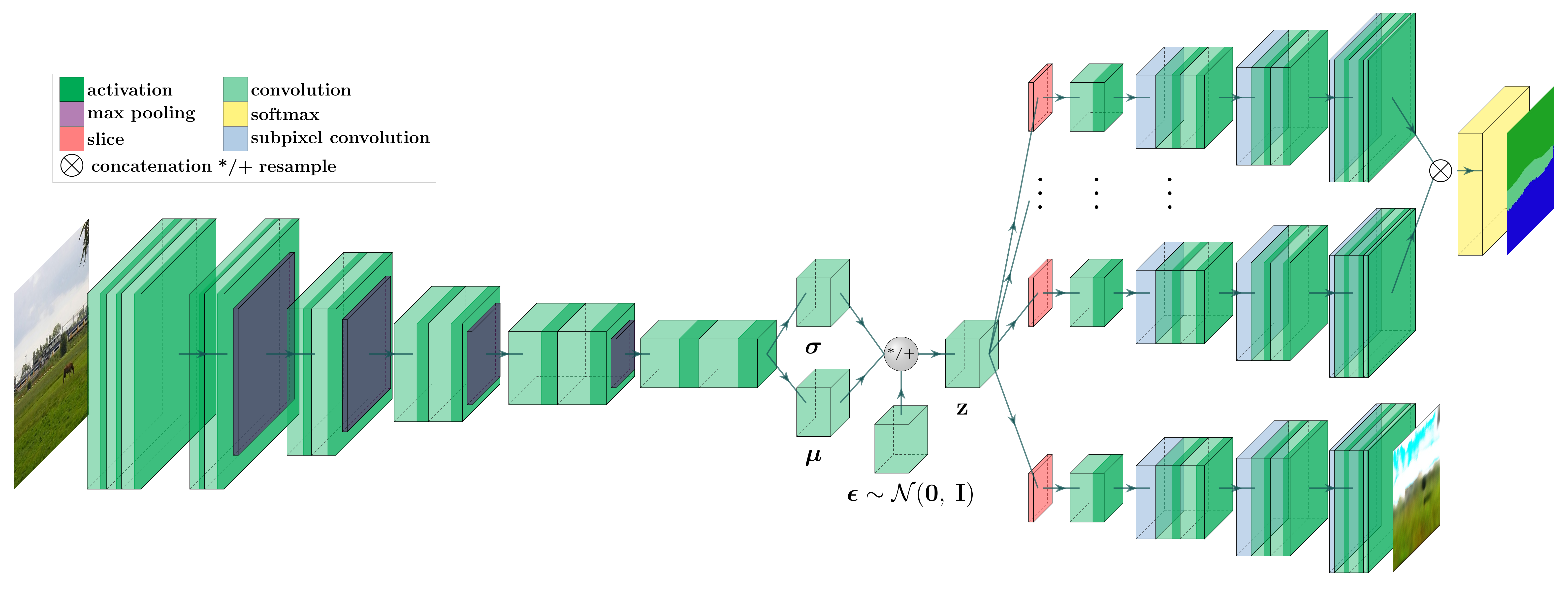}
	\caption{Our model encodes visual appearance and semantic information into separate feature maps of the latent space $\mathcal{Z}$. After the encoder, the latent variable $\mathbf{z}$ is split channel-wise into $N+1$
    local descriptors of shape $\frac{H}{16} \times \frac{W}{16} \times M$, which can also be interpreted as $M$ 
    local descriptors corresponding to that slice's decoder.
    One of the slices is dedicated to reconstructing the full-resolution RGB image, while the other $N$ are sent to the
    decoder, concatenated, then used to predict a full-resolution semantic segmentation label.
    With this design, each $N+1$ groups of local descriptors encode only information pertaining to its object class 
    or visual appearance, forcing the network to automatically place related features into the 
    corresponding feature maps of $\boldsymbol{\mu}$ and $\boldsymbol{\sigma}$.
    Note that the encoder is actually a siamese network with shared weights for the true positive images, but due to space constraints this is not depicted.
    Additionally, the decoders are not drawn to scale for the same reason.
    Best viewed in color.}
	\label{fig:arch}
\end{figure*}

We build a single custom CNN that is light-weight and amenable for real-time use in practice,
to extract all the features needed for loop closure.
As shown in Fig.~\ref{fig:arch}, the proposed network is composed of three key parts: a VAE, a semantic segmentator, and a siamese triplet embedding (siamese network is not depicted), 
while the final network for inference only consists of the encoder.
The input to our network is RGB images of height $H$ and width $W$.
Our encoder consists of a $3\times3$ {\em conv} layer followed by two residual blocks, which feed into  four blocks of $2\times${\em conv} + {\em pool}.
Finally, there are two separate $1\times1$ {\em conv} layers to calculate the latent variables $\boldsymbol \mu$ and $\boldsymbol \sigma$.
The latent variables are trained to parameterize a Gaussian distribution $\mathcal{N}(\boldsymbol{\mu}, \diag(\exp(\boldsymbol{\sigma})))$ -- where the exponential of $\boldsymbol \sigma$ is taken simply to improve numerical stability.
With this interpretation, the latent variables are both vector quantities, which are retrieved from flattening the 
$\frac{H}{16} \times \frac{W}{16} \times M(N+1)$
3D arrays they are stored in.

The latent variables are optimized to construct a standard normal distribution via a Kullback-Leibler Divergence loss:
\begin{align} \label{eq:kld}
    L_{KLD}\ \dot{=}\ & \mathcal{D}[\mathcal{N}(\boldsymbol{\mu}, \diag(\exp(\boldsymbol{\sigma})))\ ||\     \mathcal{N}(\mathbf{0}, \mathbf{I})] \notag \\
    =\ & \frac{1}{2}((\sum_i \exp(\sigma_i) - \sigma_i + \boldsymbol{\mu}^\intercal\boldsymbol{\mu} - 
    \dim(\boldsymbol{\mu})) 
\end{align}

After resampling with $\boldsymbol{\epsilon}$, which is sampled from a standard normal distribution, the latent variable $\mathbf{z} = \boldsymbol{\mu} + diag(exp(\boldsymbol \sigma))^{\frac{1}{2}}\boldsymbol{\epsilon}$ is sliced into $N+1$ groups of feature maps corresponding to visual appearance and $N$ object classes.
Optimizing with this variational objective function helps the encoder to distribute features well throughout the descriptor space, and has seen great success, for example in~\cite{Schonberger2018semantic}.

The sliced parts of $\mathbf{z}$ are fed into $N+1$ \textit{independent} decoders to decode the features corresponding to appearance and the object classes separately.
The output of the decoder for visual appearance is then sent to an RGB reconstruction loss:
\begin{align} \label{eq:rec}
    L_r = & -\sum_{h, w, c}\Big( x_{h,w,c} \log(r_{h,w,c}) \notag \\
          & \hspace{1cm} + (1 - x_{h,w,c}) \log(1-r_{h,w,c})\Big)
\end{align}
where $x_{h,w,c}$, and $r_{h,w,c}$ are the input image and reconstruction evaluated at index $(h,w,c)$, respectively.
The output of the decoders for semantic segmentation are concatenated channel-wise, and sent to a standard pixel-wise softmax cross entropy loss $L_s$ with loss weights to account for class biases.
The loss weights for each class are calculated as the inverse of the percent of all pixel labels in the dataset that contain that class, normalized to make the most prevalent class have a weight of 1.
We train our model on the COCO ``stuff'' dataset~\cite{caesar2018cvpr}.
This dataset is optimal for our case since it contains a large number of training samples, and the class definitions are focused on static objects such as buildings, walls, etc.
Rather than using the 92 stuff classes provided by COCO, we create 13 superclasses that more generally describe the semantics of a scene.
This helps to improve the model's segmentation prediction accuracy during training and lower the number of required local descriptors -- yielding a more compact embedding.
All possibly dynamic objects such as cars and people are contained in the ``other'' class.
This is to ensure that the network focuses more of its descripting power on static objects that are better to observe for loop closure.

All of the {\em conv} layers use an Exponential Linear Unit (ELU)~\cite{ClevertUH15} activation except for the layers that calculate the latent variables and the final $1\times1$ {\em conv} layers that are at the end of the decoders.
The final layers for the semantic segmentation decoders as well as the layers to calculate the latent variables have no activation, while the image reconstruction decoder has a sigmoid activation.
$2\times2$ max pooling is used in the encoder with a stride of 2 to downscale the features, and subpixel convolution~\cite{ShiCHTABRW16} is used to upscale decoder features.

\begin{figure}
    \centering
    \includegraphics[width=.9\columnwidth]{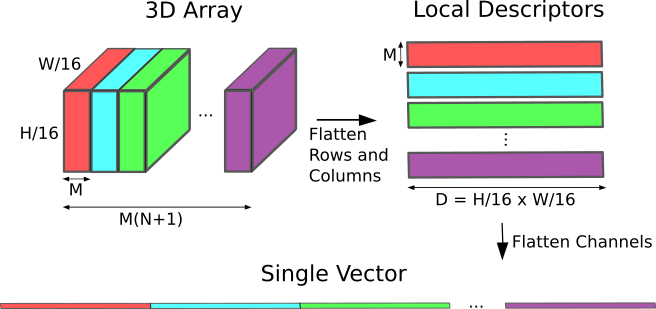}
    \caption{Here you can see a visualization of the different interpretations of our descriptor. It can be interpreted as a 3D array, 
    a set of local descriptors, or a single vector. 
    Here each color represents the data pertaining to a specific decoder.
    Note that in the form of local descriptors, each color represents 
    $M$ local descriptors of dimension $D$.
    Best viewed in color.}
    \label{fig:descr-vis}
\end{figure}

Our triplet embedding network shares weights between two encoders.
The first encoder is the one depicted in Fig.~\ref{fig:arch}, and the second is for true positive images.
Since we train on the COCO ``stuff'' dataset, true positive images are not available.
The method for extrapolating fake true positive images is discussed in the next section.
We do not require a third siamese encoder, since the hard negatives are mined from the current training batch, which assumes that all images in the COCO dataset are of different places.
Our global image descriptor is taken from the latent code $\boldsymbol \mu$, which can either be interpreted as a 3D array,
a set of $M \times (N+1)$ local descriptors of dimension $D$ pertaining to each of the network's $N+1$ decoders given convolutional features with $M$ feature maps, or a single vector of length $D \times M \times (N+1)$ 
 -- where $D=\frac{H}{16}\times\frac{W}{16}$ in our design.
 A visualization of these interpretations can be seen in Fig.~\ref{fig:descr-vis}.

With the second definition of $\boldsymbol \mu$ in mind, we first take the residual
$\boldsymbol{\mu} - \mathbf{c}$, where $\mathbf{c}$ is the channel-wise concatenation of $M\times (N+1)$ learned 
cluster centers of dimension $D$, which are randomly initialized with a Gaussian distribution and optimized to minimize the triplet embedding loss.
The residuals are then intra-normalized
in the same manner as NetVLAD~\cite{Arandjelovic16} using the $\ell_2$ norm across channels to reduce bursts in the descriptors. 
Then, with the first definition of $\boldsymbol \mu$ in mind, we  normalize the entire
 descriptor to allow for cosine similarity calculation by inner product.

Given the normalized whole image descriptors for a database image $\mathbf{d}_d$, true positive image $\mathbf{d}_p$, and negative image $\mathbf{d}_n$, the triplet embedding objective function is defined by the following hinge loss:
\begin{align} \label{eq:triplet}
    L_t = \max(0, \mathbf{d}_d^\intercal(\mathbf{d}_n - \mathbf{d}_p) + m)
\end{align}
where $m$ is the margin hyperparameter.
This loss function forces the network to learn to separate the similarity between positive and negative images by the margin $m$.
By minimizing this objective, the cosine similarity between the database and negative image representation is minimized, while the similarity between the database and positive representations are maximized.
If the training example can separate these two similarity scores by $m$, then no loss is incurred.

\subsection{Network Training}

The model is trained with the Adam optimizer \cite{KingmaB14}.
The overall objective function is defined by:
\begin{equation} \label{eq:objective}
    L = \lambda_0 L_{KLD} + \lambda_1 L_{r} +
            \lambda_2 L_s + \lambda_3 L_t
\end{equation}
where $\lambda_i$ are for loss scaling.
We utilize two GTX 1080-Ti GPUs to train the model, each holding a batch size of $N_b$.
The model is trained  until the testing accuracy diverges, which is determined by the AUC for the Campus Loop dataset~\cite{Merrill2018RSS}.

\subsubsection{True Positive Extrapolation}
Since the COCO dataset does not contain true positive images, we must create our own.
We use the method described in~\cite{Merrill2018RSS} to randomly warp training images using homographies, which can emulate camera motion to a certain degree.
The difference here is that we warp the images on the fly so that the same homographic warp is not seen twice in training, which adds some computation, but is still fast since we compute the warps in parallel batches across all GPUs used to train.
Additionally, we randomly darken the images that have a mean intensity higher than a threshold $\tau$ to account for the lack of night-time images in the COCO dataset.
We also randomly left-right flip the warped and darkened images in order to account for the limitations in the amount we can warp images with homographies.
The result is a fake true positive image that is not as valuable as a real one would be, but meets our needs to achieve viewpoint invariance.

\subsection{Inference}

\subsubsection{Keypoint Extraction}

While global image descriptors are useful for image retrieval via nearest-neighbor searches, they require thresholding to actually determine a match.
To combat this issue, we opt to extract keypoints from maximally-activated regions of the low-level {\em conv5} layer of our network in a similar manner to~\cite{garg2018lost}.
While the number of convolutional feature maps used by~\cite{garg2018lost} far outnumbers that of possible keypoint locations, we have the opposite issue.
Our {\em conv5} layer is at full resolution, and has 32 feature maps.
To extract a meaningful number of keypoints, we simply take the keypoints to be the maximally activated regions over $H/N_w\times W/N_w$ windows of each feature map. 
Clearly varying $N_w$ will return a variable number of keypoints, and is a hyperparameter of our system.
Once initial keypoints are extracted, duplicates are removed.
We found that there are a relatively small number of duplicates, and the initial keypoints are well-distributed over the image plane due to the windowing scheme.

After the set of keypoints is computed, we extract our novel keypoint descriptors.
Garg et. al~\cite{garg2018lost} use the normalized vector of feature maps evaluated at a single pixel location to construct a keypoint descriptor amenable for comparison on the unit hypersphere, 
and filter the keypoint matches using the semantic segmentation prediction from their system.
However, dense semantic segmentation is costly, so we opt to avoid that task and simply construct a better keypoint descriptor from the {\em conv5} features.
Inspired by hand-crafted keypoint descriptors such as BRIEF~\cite{BRIEF}, which are computed from a series of binary tests over points surrounding an interest point, where the image intensity is directly compared, we perform a similar windowed operation on the convolutional feature maps.
In contrast to a single intensity value at a $(u,v)$ coordinate, we have a pre-computed feature vector of length 32, so instead of comparing the image intensity, we can directly compare the activations.
To this end, we take the residual of each of the feature vectors surrounding the interest point with respect to the interest point's feature vector in a $3\times 3$ window.
By concatenating these residuals, we obtain a 256-dimensional keypoint descriptor that
 is more discriminative than the original 32-dimensional feature vector from the interest point.
These descriptors are directly compared using the Euclidean distance metric during keypoint matching.
To match keypoints between database and query images, we simply use a K-nearest neighbor search with $K=2$, and use the traditional ratio test with ratio $r$ to determine a valid match.
Our keypoint matching method is not as robust as the best hand-crafted features, but it serves as a good geometric check for loop closure candidates.

\subsubsection{Loop Closure Detection}

\begin{figure} [t]
    \centering
	\includegraphics[width=.99\columnwidth]{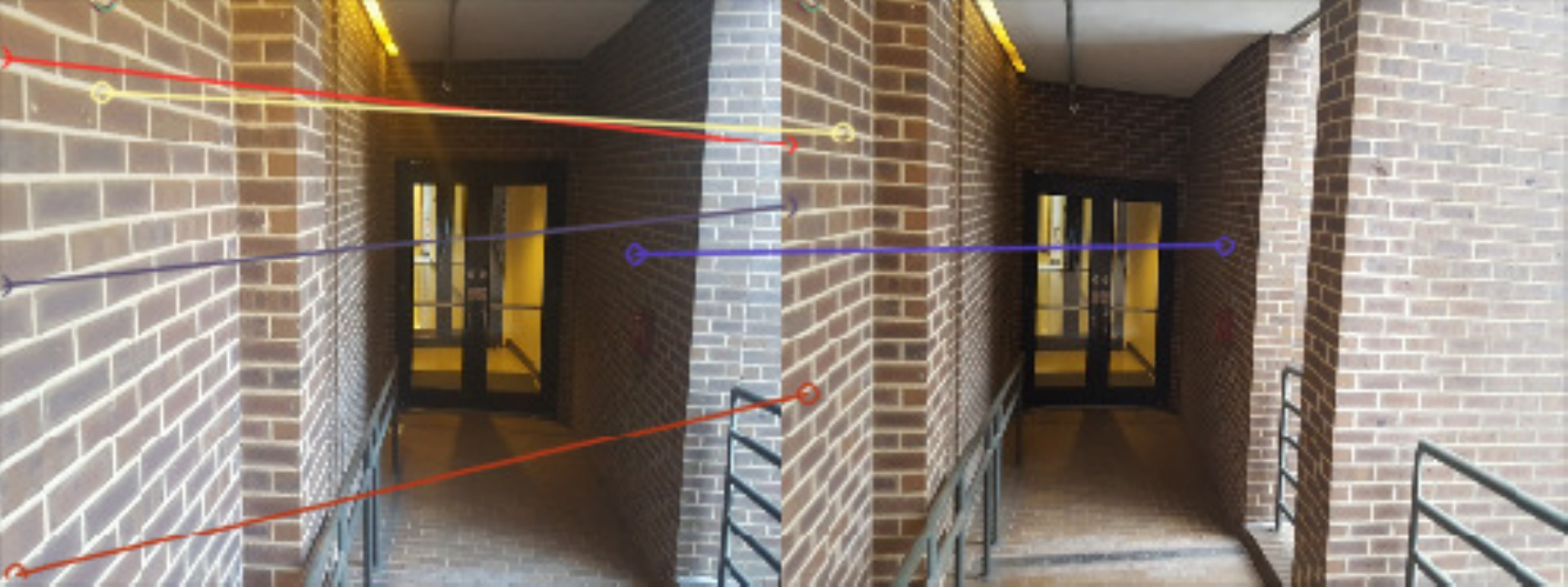}
    \caption{An example of matched keypoint inliers after performing RANSAC.
            Clearly these keypoints are not abundant enough for use in pose estimation,
            but they serve as a good geometric check for loop closure detection.
             }
    \label{fig:kp}
\end{figure}

By combining the fast yet discriminative power of our global image descriptor with the geometric capabilities of our keypoints, we are able to accurately close loops without the need for thesholding.
To determine a loop closure, we first perform a K-nearest neighbor search with $K=7$ over the database of global image descriptors.
We then filter the K candidates by matching keypoints as described above. 
Candidates are rejected that do not have enough valid matches to estimate a fundamental matrix with the RANSAC algorithm, which requires a minimum of 8 matches.
An example of final inlier matches after performing RANSAC can be seen in Fig.~\ref{fig:kp}.
From there, the final candidate is taken as the one with a valid number of matching keypoints and also the highest global descriptor similarity score.
If a valid fundamental matrix can be computed with the matching keypoints, a loop is considered detected; otherwise the candidate is rejected.
In practical applications, this approach is further robustified by ensuring that multiple sequential matches are valid.

\begin{table} [t]
    \centering
    \caption{List of Hyperparameters}
    \begin{tabular}{@{\extracolsep{4pt}}lllllllll} 
        \toprule
         $H$ & $W$ & $M$ & $N_b$ & $m$ & $\tau$ & $r$ & $N_w$  \\
        \midrule
         192 & 256 &  4  &  12   & 0.5 &  0.2   & 0.7 &   4 \\
        \bottomrule
    \end{tabular}
    
    \vspace{.5cm}
    
    \begin{tabular}{@{\extracolsep{4pt}}llll} 
        \toprule
         $\lambda_0$ & $\lambda_1$ & $\lambda_2$ & $\lambda_3$ \\
        \midrule
         $10^{-4}$ & $10^{-4}$ & 1.0 & 1.0 \\
        \bottomrule
    \end{tabular}
    \label{table:hp}
\end{table}


\section{Experimental Results} \label{sec:experiments}


We construct and train our network entirely with TensorFlow~\cite{tensorflow}.
The Adam optimizer has the default TensorFlow parameters.
The homographic estimation and warping is done with custom TensorFlow code, which is available in the source code repository linked in this work.
The keypoint extraction and keypoint descriptors are computed after retrieving the {\em conv5} data from the GPU, and we use the OpenCV~\cite{opencv_library} bruteforce matcher to match keypoints as well as their RANSAC algorithm.
The source code is written in Python.
A list of all hyperparameters used in our system can be seen in Table~\ref{table:hp}.
The local descriptor dimension $D$ is $H/16 \times W/16 = 192$ due to the encoder structure, making the total global descriptor dimension 10,752.

To validate the performance of the proposed CALC2.0 network, we compare it against the state-of-the-art method NetVLAD~\cite{Arandjelovic16}.
In particular, we evaluate on three datasets, all of which are exactly the same as used in our prior work of CALC~\cite{Merrill2018RSS}
-- although we do not apply any precision-recall interpolation since it hides some fine details in some of our experiments that determine the better algorithm.
Performance metrics used are the Area Under Curve (AUC) and the highest recall rate at 1.0 precision denoted by $r$ in the plots (which is different from our ratio test hyperparameter $r$).
Both of these values are included in the plot legends.

In the results presented below, 
our proposed methods are denoted as CALC2 (dashed blue line) and G-CALC2 (solid blue line),
while NetVLAD~\cite{Arandjelovic16} is depicted in a dotted-dashed red line.
Additionally, we compare against our previous method CALC, which is depicted in solid green.
Note that the AUC values are slightly different for CALC here than in~\cite{Merrill2018RSS} since we do not interpolate the precision recall curves.
For CALC2, we simply perform a nearest neighbor search with the global image descriptor, while G-CALC2 (Geometric CALC2) uses the geometric keypoint checks.
Due to its design, G-CALC2 may not always return a match.
If this is the case, the similarity score is set to the minimum -1, and the query is deemed incorrect.
Note however that it is still possible to achieve perfect recall with this design even though it may return some false negatives.
No sequence information is used in these precision-recall experiments since temporal consistency logic is a common practice and can only improve the results.
Additionally, none of the methods in these experiments have seen any of the testing data during training, making for a very fair comparison.

\subsection{Campus Loop Dataset}

\begin{figure}
    \centering
	\includegraphics[width=.99\columnwidth]{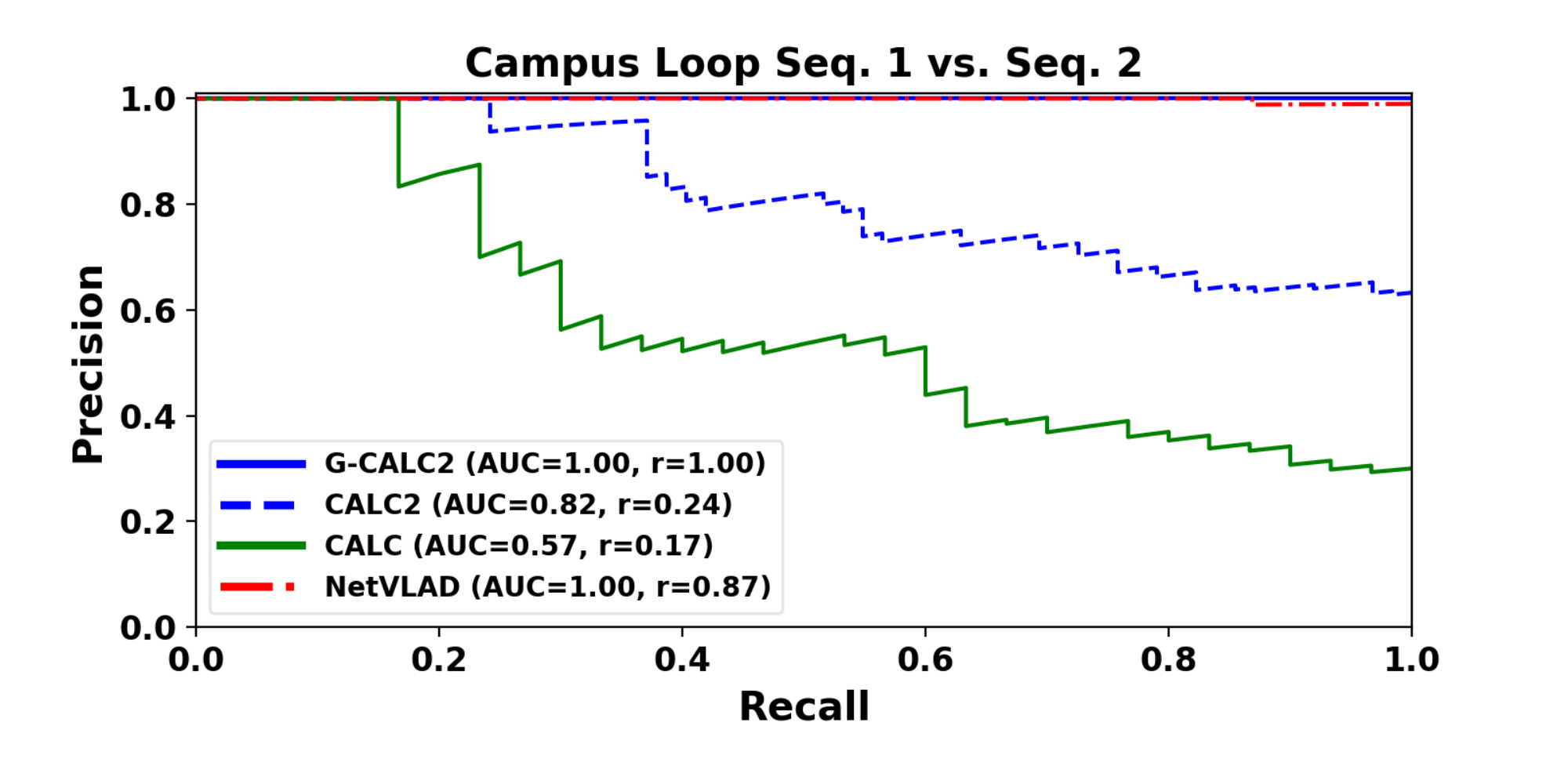}
    \caption{For the Campus Loop dataset, our G-CALC2 method outperforms NetVLAD, returning only correct matches.
             }
    \label{fig:cl}
\end{figure}

\begin{figure} 
    \centering
	\includegraphics[width=.31\columnwidth]{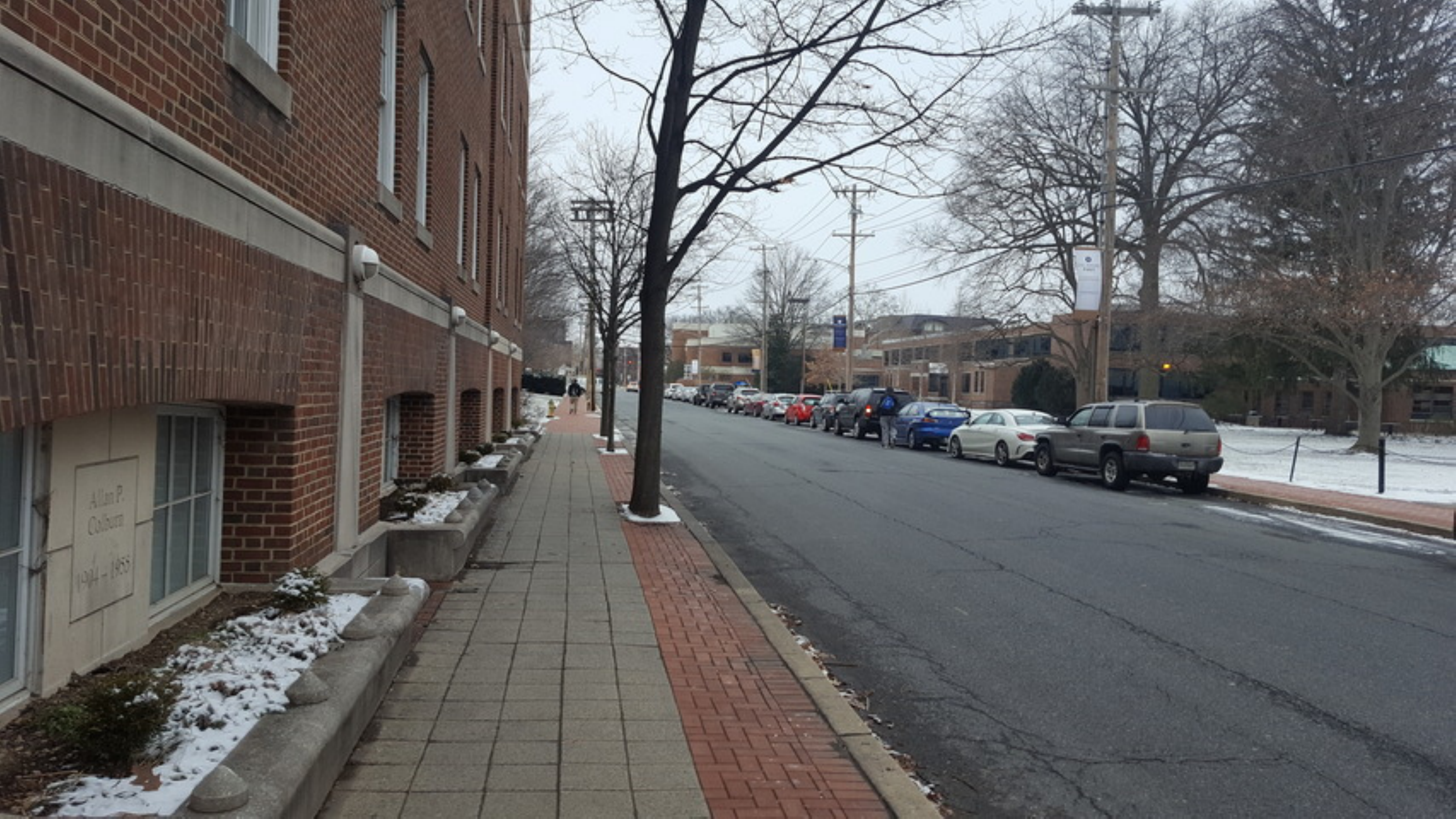}
	\includegraphics[width=.31\columnwidth]{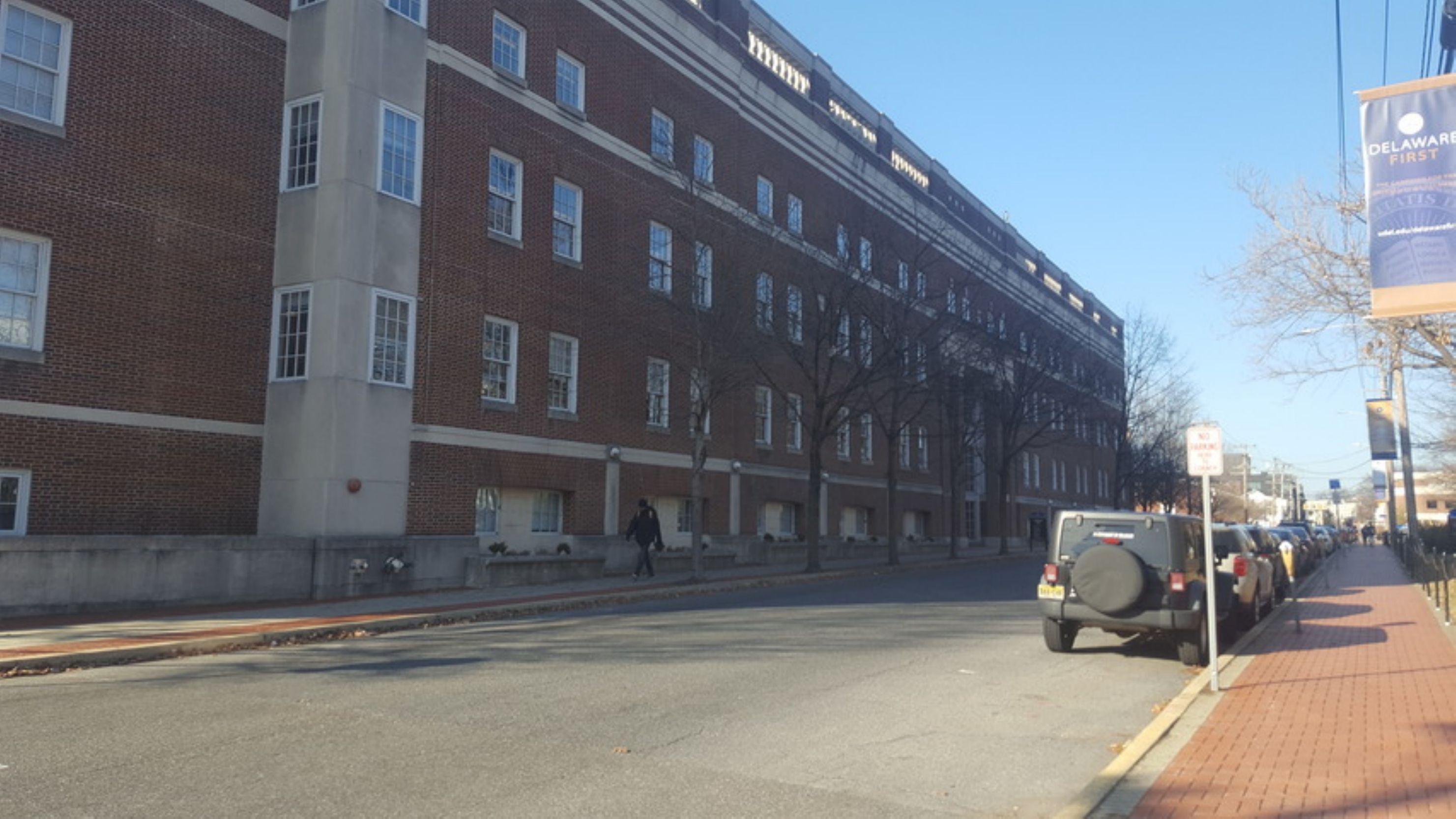}
	\includegraphics[width=.31\columnwidth]{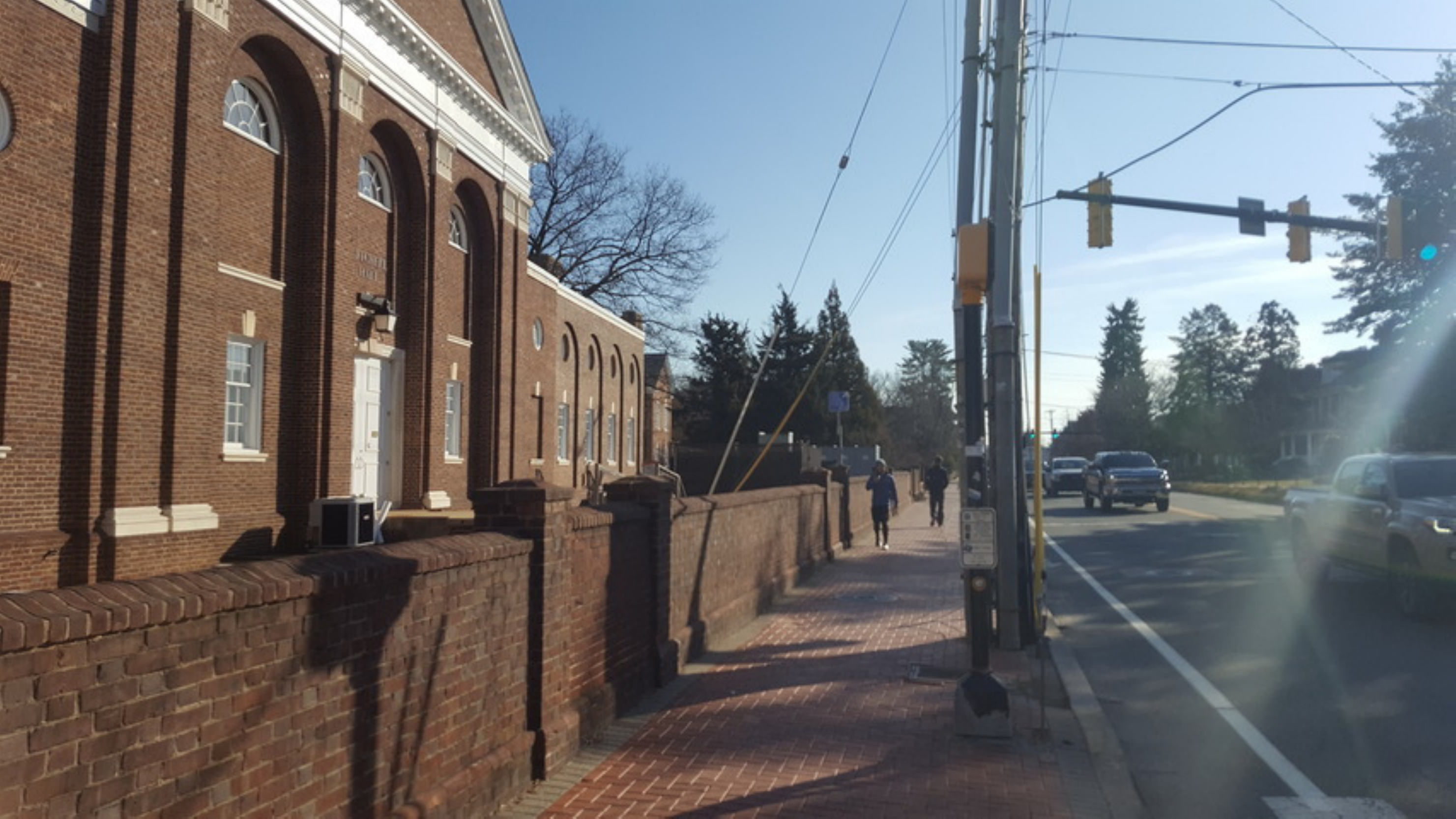}
	\includegraphics[width=.99\columnwidth]{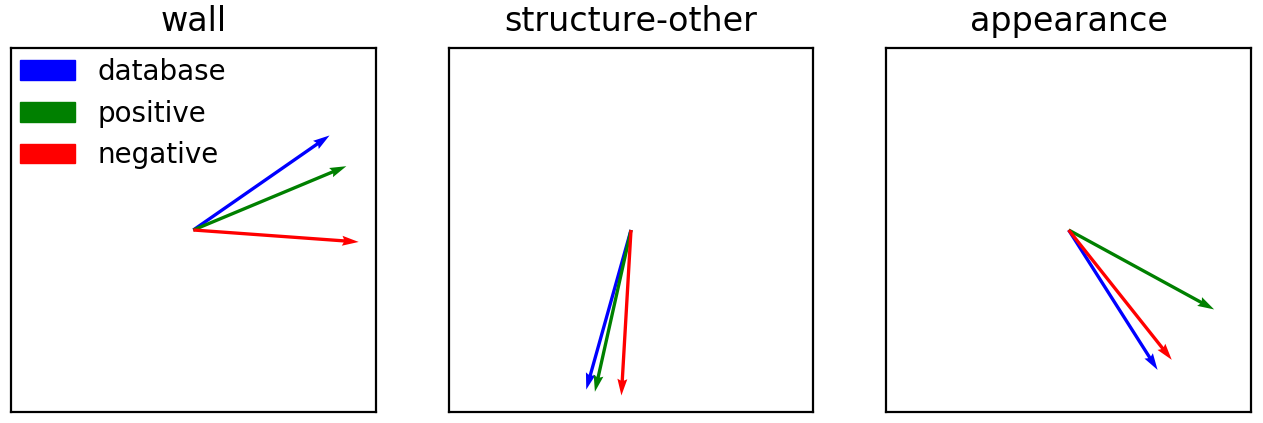}
    \caption{Visual appearance can be misleading, but by utilizing semantic information, our descriptor can remain discriminative.
            Descriptors are computed for the images from the Campus Loop dataset shown above.
            There is a database image (top left), true positive query image (top center) and true negative (top right).
            We compute two principal components of the local normalized residual descriptors corresponding 
            to the ``wall'' class (bottom left), the ``structure-other'' class (bottom center) and visual appearance (bottom right).
            The database descriptor is shown in blue, the true positive in green, and the true negative in red.
            As you can see, the principal components corresponding to the visual appearance for the database image are closer on the unit circle to the negative descriptor than the positive, while the semantic descriptors successfully map closer to the positive.
            This is most likely due to the fact that the true negative and the database image are visually more similar, but the database image and the true positive both lack a wall as well as the light post that would be classified as ``structure-other''.
            Best viewed in color.}
    \label{fig:pc}
\end{figure}

The Campus Loop dataset is a challenging dataset that was presented in~\cite{Merrill2018RSS}.
It consists of two sets of 100 frames from around the University of Delaware campus.
The dataset contains large viewpoint changes as well as luminocity and appearance variations from clouds and the presence/absence of snow.
We first demonstrate the discriminative power of our local image descriptors extracted from the full normalized descriptor.
We take two principal components of the descriptors for the ``wall'' and ``structure-other'' classes in our custom super classes, where the PCA whitening matrices are trained separately for each class on the entire Campus Loop Dataset.
It can be seen from Fig.~\ref{fig:pc} that the semantic information maps more closely in descriptor space than the visual information, 
since the database and true negative image are more visually similar yet semantically different.
This analysis shows the benefit of including semantic information in an image descriptor, since our system successfully matched these images in the following experiment.

More importantly, 
Fig.~\ref{fig:cl} depicts the loop closure results on this dataset,
which clearly shows that our G-CALC2 method achieves perfect recall, outperforming NetVLAD, by a small margin.
Note that this does not mean that G-CALC2 always returned the correct image, as it did have some missed detections; however, when it did return a match, it was always correct.
This makes G-CALC2 a more conservative algorithm than a thresholded nearest neighbor search algorith, but in practice it is typically better to miss detections than to return false positives.
CALC performs significantly worse than the other methods here -- likely due to the fact that the Campus Loop dataset contains viewpoint variations that are larger than the homograghic warping used by that method can emulate.
It is important to note that while NetVLAD is very competitive to our system, it utilizes full-resolution images, and takes on the order of {\em hundreds} of milliseconds to compute an image representation in our experiments -- depending on the image resolution.
On the contrary, our system uses down-sampled images of $192\times256$ resolution, and takes only about {\em 3-5} milliseconds to compute the global image descriptors on our desktop computer, 
thus making our system amenable for real-time performance.


\subsection{Nordland Dataset}

\begin{figure}
    \centering
	\includegraphics[width=.99\columnwidth]{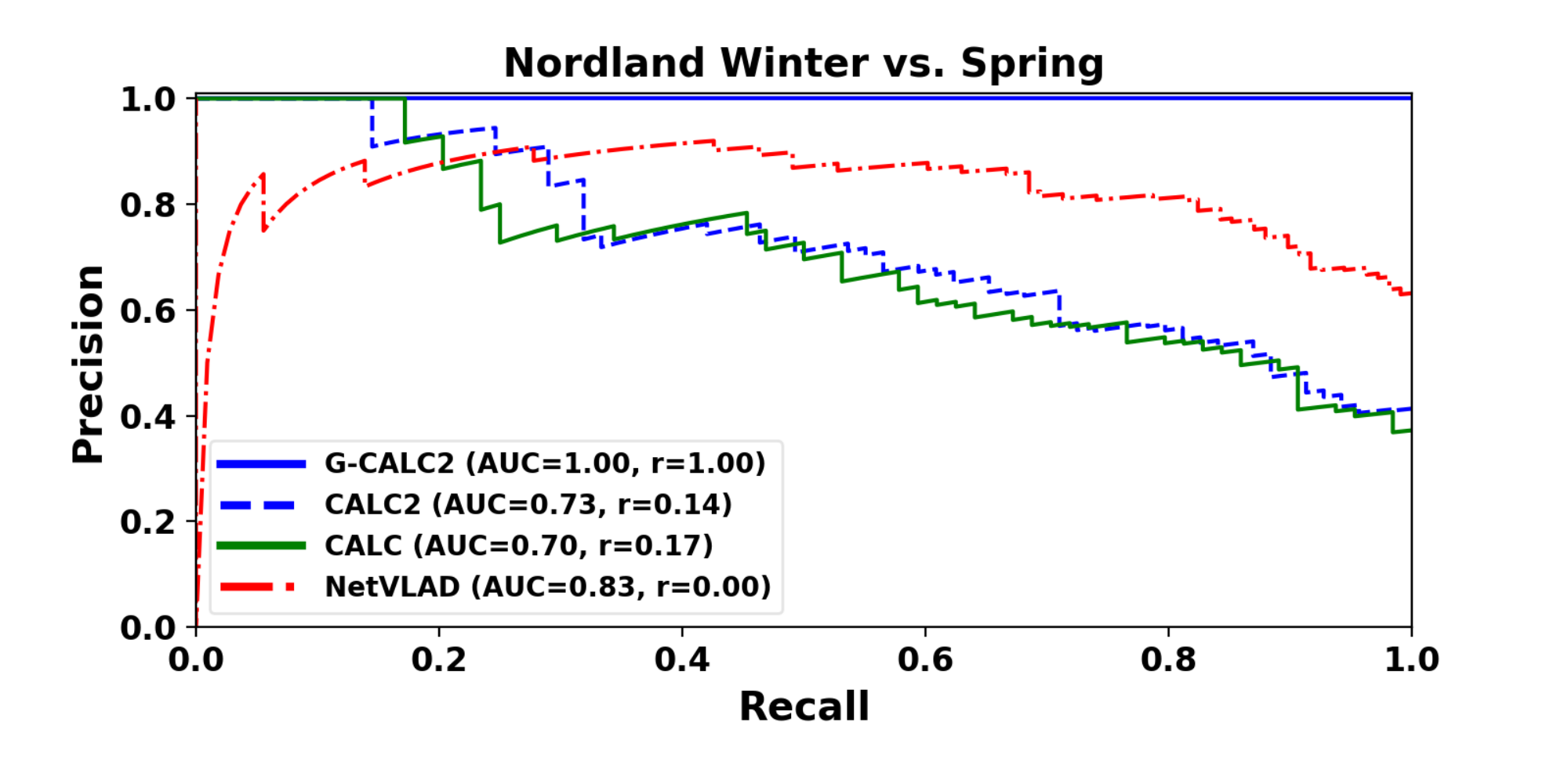}
    \caption{In one the most challenging of the Nordland sequence combination, winter and spring, our method outperforms NetVLAD -- only returning correct matches.
            NetVLAD does not have perfect precision at any recall rate, despite utilizing the full $1920\times1080$ resolution images.
             }
    \label{fig:nordland}
\end{figure}

The Nordland dataset was collected over different seasons in the same locations from a moving train.
Here we use the same subset of this dataset as in~\cite{Merrill2018RSS} -- the winter vs. spring sequence.
These two sequences exhibit large appearance differences due to the presence of snow in the winter images, but, in contrast to the Campus Loop dataset, contain little to no viewpoint changes since the images are time synchronized and taken from the same point of view on the train.
The results from this experiment can be seen in Fig.~\ref{fig:nordland},
which also shows G-CALC2 achieves best recall, while NetVLAD fails to achieve any recall -- although it has a fairly high AUC.
CALC and CALC2 are very comparable in performance here, where CALC2 has a higher AUC, but CALC has a higher recall rate.
This may be due to the fact that CALC uses grayscale images, so the visual differences between images with and without snow are not as clear as with the RGB images used by our current system.
On the other hand, many of the Nordland images look very similar, so the semantic descriptors we currently deploy can combat this issue more so than CALC -- leading to an interesting trade-off.

\subsection{Gardens Point Dataset}

\begin{figure}
    \centering
	\includegraphics[width=.99\columnwidth]{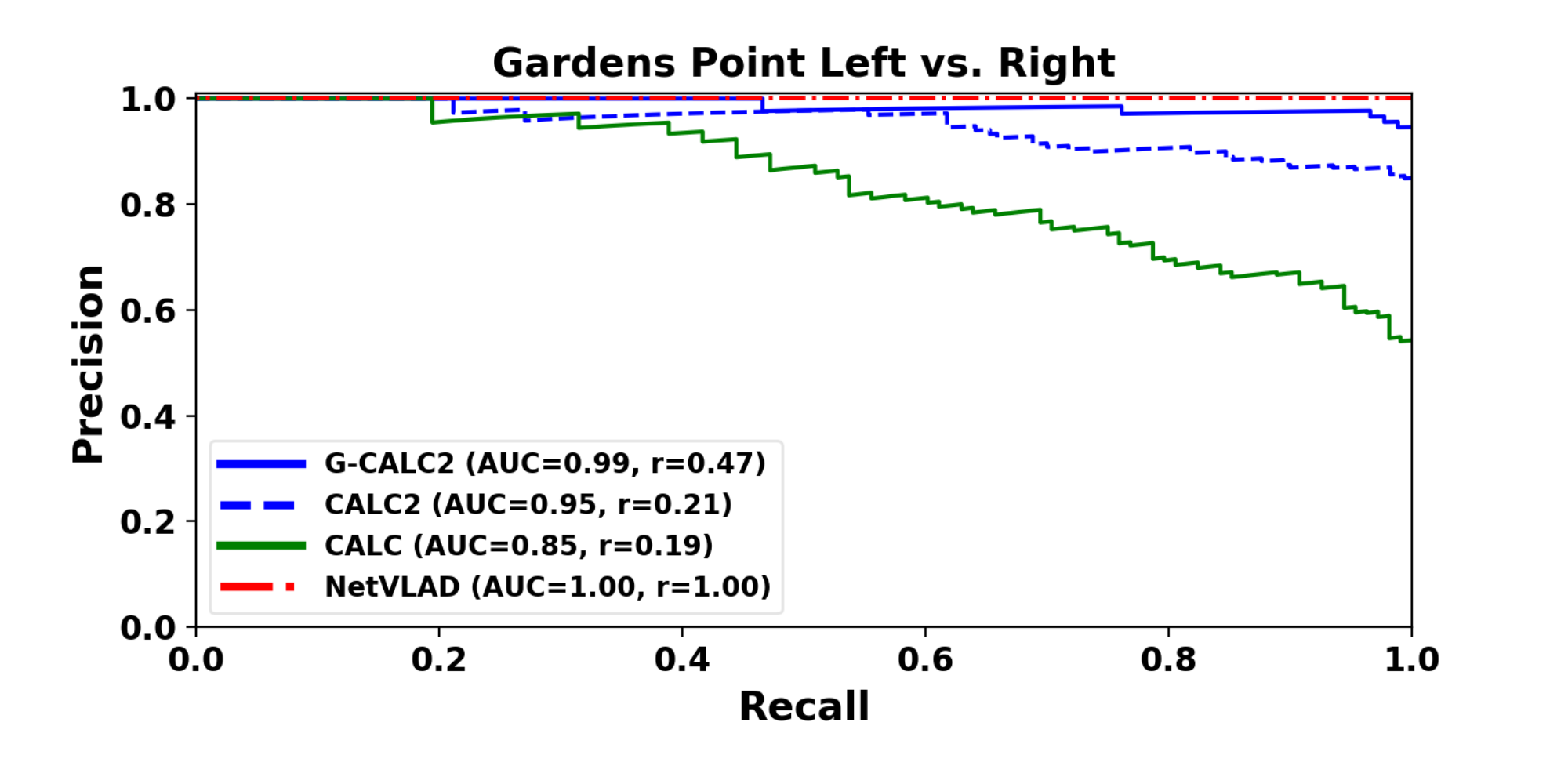}
    \caption{While our method is competitive in the Gardens Point dataset, NetVLAD outperforms ours.
             }
    \label{fig:gp}
\end{figure}

The Gardens Point dataset consists of two daytime traversals through a university campus,
which incorporates moderate viewpoint changes as well as many dynamic objects -- typically students.
Unlike the Nordland and Campus Loop datasets, the Gardens point dataset does not exhibit weather changes.
The results from this experiment are shown in Fig.~\ref{fig:gp}.
In this case, NetVLAD achieves perfect recall while our method falls slightly short.
However, both the CALC2 and G-CALC2 methods are highly competitive in this case, with nearly perfect AUC.
CALC falls short of all the algorithms here.
It is important to note that while CALC may not be the most accurate of these systems, it is the fastest, so clearly there are trade-offs to consider.

\subsection{Loop Closure In a Practical Environment}

\begin{figure} [t]
    \centering
	\includegraphics[width=.99\columnwidth]{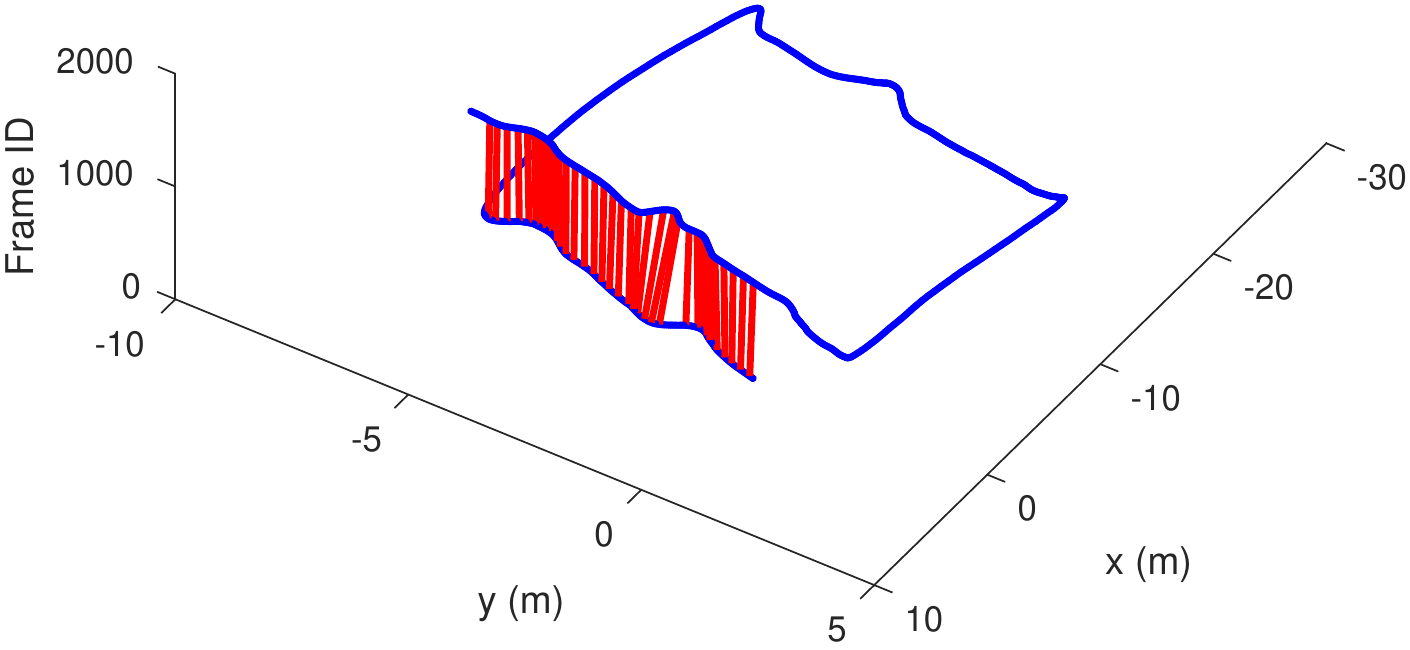}
    \caption{We demonstrate the performance of our system on the KITTI odometry dataset sequence 6.
            The position is on the horizontal plane while the frame index is on the vertical axis.
            It is clear that our method can correctly close loops in a practical application.
             }
    \label{fig:kitti}
\end{figure}

Here we demonstrate the performance of our G-CALC system in a real-world loop closure scenario,
 using the KITTI odometry dataset~\cite{Geiger2012CVPR} sequence 6 color images.
The images are resized to our $192\times 256$ resolution with no cropping or preprocessing.
We fill a database with images as they are received, and wait until we have 200 frames to begin closing loops, and in this experiment we process every frame.
It is important to note that frame 200 occurs before the first turn in this trajectory, so the system is looking for matches throughout most of the frames.
In this experiment, we use the same hyperparameters as in Table~\ref{table:hp}, but add a simple temporal consistency check -- that is, if eleven sequential frames are matched to a database images within a small window of frame IDs, a loop is considered detected.
The results of this experiment can be seen in Fig.~\ref{fig:kitti}.
Clearly there are no false positives in this experiment, and there are essentially no missed detections.
By using every frame, we get a large number of loop closures.
This can be sparsified by only detecting on keyframes, or refuting loop closures if one happened too recently.
Our system can perform the nearest neighbor searches and geometric checks at over 20 Hz by the end of the dataset, even with our current inefficient Python implementation.
The speed can potentially be further increased by removing detected loop descriptors over a window from the database -- lowering the number of descriptors to search.
However, we do not do that here in order to observe the accuracy performance of the system better.

\section{Conclusions and Future Work} \label{sec:conclusions}

We have presented a novel deep loop closure system that incorporates semantic, appearance, and geometric information to detect loops.
The system is a combination of a VAE, semantic segmentator and triplet embedding network, and learns to automatically bin information into its local descriptors for semantic classes and visual appearance.
The system is competitive with the state-of-the-art NetVLAD network, while not requiring any user-defined thresholds to determine loop closures and having significantly better efficiency.
%
%
In the future, as the current implementation is in Python, we will release a C++ implementation with the source code in order to improve efficiency and allow seamless integration into real-time SLAM systems.
We will additionally work to improve the keypoint capabilities of our network -- observing that the current keypoints serve as a good geometric check, but too many of them get filtered out to actually be used for pose estimation.

\printbibliography
\end{document}